\documentclass[letterpaper, 10 pt, conference]{ieeeconf}  

\IEEEoverridecommandlockouts                              

\usepackage{amsmath}
\usepackage{algorithm}
\usepackage{algpseudocode}
\usepackage{amssymb}
\usepackage{gensymb}
\usepackage{amsmath}
\usepackage{hyperref}
\usepackage{graphicx}
\usepackage{caption}
\usepackage{subcaption}
\usepackage{pifont}
\usepackage{xcolor}
\graphicspath{ {./img/} }
\usepackage{color}
\usepackage{mdframed}
\usepackage[table]{xcolor}
\usepackage{url}
\usepackage{tcolorbox}
\usepackage{pifont}
\usepackage{xcolor}
\usepackage{leftidx}
\usepackage[noadjust]{cite}
\usepackage{booktabs}
\usepackage{soul}
\usepackage{array}
\usepackage{xcolor}
\usepackage{tikz}
\usepackage{multirow}
\usetikzlibrary{positioning}
\usepackage[top=0.75in, bottom=0.75in, left=0.75in, right=0.75in]{geometry}
\usepackage{lettrine}
\usepackage{multirow}
\usepackage{booktabs}
\usepackage{tikz}
\usepackage{float}

\newcommand{\mh}[1]{\textcolor{black}{#1}}
\title{\LARGE \bf
MAPL: Multi-Objective Preference Learning for Robot Locomotion}

\author{Xiyue Chen$^{1}$, Muhan Lin$^{2}$, Shuyang Shi$^{3}$ and Joseph Campbell$^{2}$
\thanks{$^{1}$ XC is with the Department of Electrical and Computer
Engineering, Purdue University, West Lafayette, IN 47906, USA
        {\tt\small chen4513@purdue.edu}}%
\thanks{$^{2}$ ML and JC is with the Department of Computer Science, Purdue University, West Lafayette, IN 47906, USA
        {\tt\small {lin2265, joecamp}@purdue.edu}}%
\thanks{$^{3}$ independent researcher }
}

\begin{document}

\maketitle
\thispagestyle{empty}
\pagestyle{empty}

\begin{abstract}

Reward design remains a major bottleneck in reinforcement learning for robot locomotion, where successful policies often depend on carefully tuned, task-specific reward functions. 
Preference-based reinforcement learning offers an alternative, but existing LLM-based methods typically ask for a single overall judgment between behaviors, making it difficult to capture the multiple competing objectives that underlie high-quality locomotion.
We present Multi-Objective AI-Informed Preference Learning (MAPL), a framework that learns locomotion rewards from high-level natural language objectives rather than manually engineered reward equations.
MAPL prompts a large language model to compare trajectories independently along semantically meaningful criteria, using generic language descriptions that are terrain-invariant and require little domain expertise.
These objective-wise preferences are used to train a multi-head preference scoring model, whose outputs are aggregated to form a scalar reward for policy optimization.
Across four quadruped locomotion environments, MAPL trains policies using only LLM-generated preferences and achieves performance comparable to or better than expert-designed rewards, while eliminating task-specific reward engineering.

\end{abstract}
\section{Introduction}
\label{sec:introduction}

Reinforcement learning (RL) has led to major advances in legged locomotion, enabling quadruped robots to learn agile and robust behaviors across challenging terrain~\cite{kumar2021rma,lee2020learning,margolis2023walk,margolis2024rapid}.
In practice, however, these successes still depend heavily on carefully engineered reward functions.
Designing such rewards is often tedious and brittle: practitioners must manually balance many competing objectives, such as command tracking and stability, and small misspecifications can lead to poor local optima or reward hacking~\cite{skalse2022defining}.
As a result, reward design remains one of the largest practical barriers to applying RL for complex robot locomotion.

Preference-based learning (PbRL) offers an alternative less reliant on domain-knowledge.
Rather than requiring the designer to fully specify a reward function, human judgments are used to identify preferential behaviors~\cite{christiano2017deep,park2022surf,lee2021pebble}.
Recent work replaces human annotators with LLMs, using them to either directly generate reward code \cite{ma2023eureka, heng2025boosting, zeng2024learning} or to produce pair-wise rankings over behaviors which are used to train reward a model~\cite{wang2022skill,wang2025primt,wang2024rl}.
This is a promising direction as it reduces human effort and opens the possibility of specifying desired behavior in natural language.

However, existing LLM-based preference learning methods typically require the model to produce a single overall ranking between behaviors.
This makes it difficult to elicit reliable preferences for locomotion, since the LLM must implicitly trade off several distinct behavioral criteria within a single judgment.
In this work, we show that LLM-based reward learning becomes substantially more effective when these criteria are ranked and modeled separately.

We propose \textbf{\underline{M}}ulti-Objective \textbf{\underline{A}}I-Informed \textbf{\underline{P}}reference \textbf{\underline{L}}earning (MAPL), \textbf{a framework that learns rewards from LLM feedback over multiple, decomposed objectives.}
MAPL queries an LLM to rank trajectories independently according to three objectives: velocity tracking, smoothness, and stability.
Crucially, these objectives are specified entirely in terms of \textit{high-level natural language}.
The resulting rankings are then used to train a model to infer rewards for each objective independently, after which they are linearly combined to yield a single scalar reward signal.
This decomposition allows the LLM to evaluate each behavioral criterion separately, rather than requiring a single global judgment that implicitly mixes several trade-offs at once.
In practice, this produces more reliable supervision and more optimal downstream policies, especially when the LLM is more accurate on some criteria than others.
We evaluate our method on quadruped locomotion across four environments: flat ground, uneven terrain, stairs, and obstacle traversal.
Our results show that MAPL can train policies without \textit{any} environment reward, using only the reward model learned from LLM-generated preference rankings, and achieves performance equal to or better than policies trained with expert-designed rewards.
Moreover, a \textit{single generic prompt} is used to train optimal policies across all four terrain types, completely eliminating the task-specific reward engineering and domain expertise normally required. 
These results suggest that LLM-based reward learning can move beyond coarse preference supervision and become a practical tool for scalable locomotion training.

\begin{figure}[]
    \centering
    \includegraphics[width=1\linewidth]{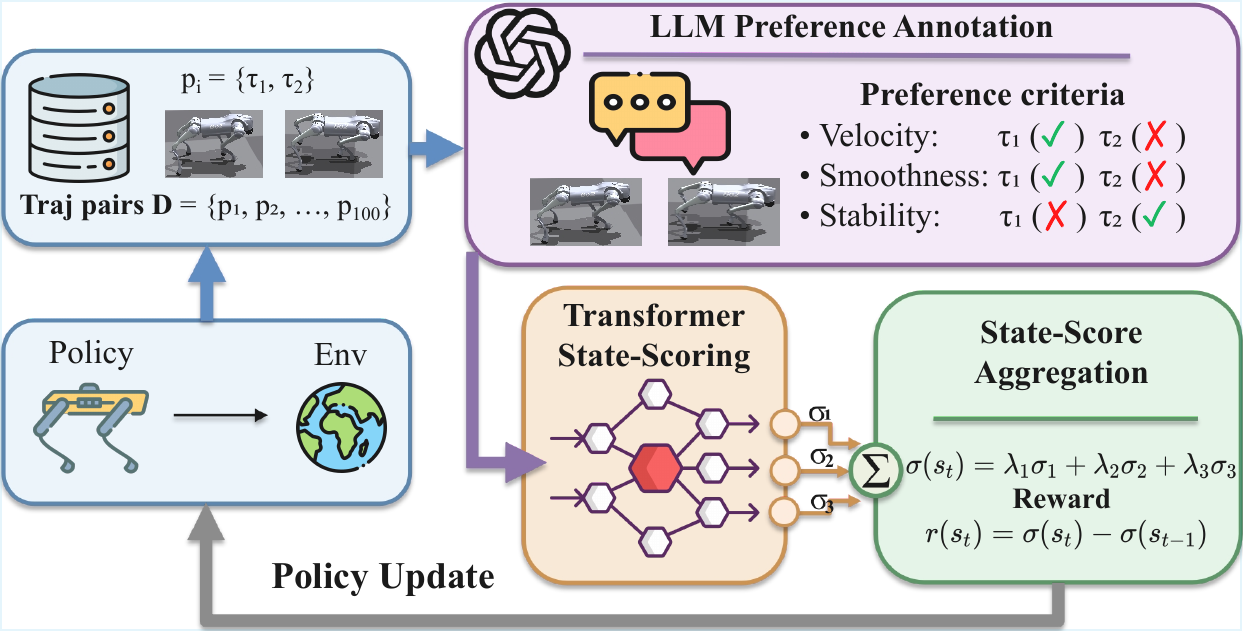}
    \caption{
    MAPL consists of two iterative steps. \textbf{Scoring Model Training:} (1) Sample trajectory pairs from replay buffer. (2) Query LLM to obtain multi-objective pairwise preferences. (3) Train a transformer-based preference scoring model for each objective.
    \textbf{Policy Training:} (1) Sample state-action pairs by rolling out policy. (2) Compute scores for each state using the preference scoring model. (3) Aggregate scores to compute the potential-difference reward. (4) Update the policy using state-action-reward triplets.
}
\label{fig:overview}
\vspace{-2ex}
\vskip -0.1in
\end{figure}
\section{RELATED WORK}
\label{sec:RELATED WORK}
\subsection{Reward Design for Legged Robot}

RL enables robots to acquire robust and agile behaviors through trial-and-error optimization guided by reward functions \cite{margolis2023walk, lee2020learning, margolis2024rapid, kumar2021rma, zhou2026learning}.
However, the reliance on expert prior knowledge and carefully engineered reward terms 
significantly limit the efficiency of RL in robot learning. Recent studies have explored leveraging LLMs to automatically generate reward functions for robot tasks~\cite{ma2023eureka, zeng2024learning, heng2025boosting}.
Nevertheless, \mh{these methods need to deal with LLM hallucination and its limited concepts about numerical reward values. Therefore,} these approaches still require \mh{iterative reward function revision by LLMs and} substantial manual tuning to achieve reliable performance in complex robot tasks.

\subsection{Preference-based RL in Robot Learning}
\mh{Preference-based RL} aims to alleviate the challenges of hand-crafted reward design by learning reward functions from human preferences over pairwise trajectory comparisons \cite{wilson2012bayesian, christiano2017deep, lee2023rlaif, wang2025primt}. Rather than specifying explicit reward terms, \mh{RL from human feedback (RLHF) trains parametrized reward} functions that better align with human notions of task success \cite{christiano2017deep}. While preference labels require extensive human efforts, more recent work, \mh{represented by RL from AI feedback (RLAIF),} replaces human annotators with pretrained models to provide scalable, automated feedback. \cite{bai2022constitutional, yu2025rlaif, li2023silkie} \mh{Some works start to apply this framework to robot learning. For robot manipulation tasks, PrefCLM and RL-SaLLM-F \cite{tu2024online, wang2025prefclm} prompt LLMs to generate preference over state–action trajectory pairs, while RL-VLM-F \cite{wang2024rl, venkataraman2025real} employs vision–language models (VLMs) to provide preferences over visual observations of states}.

\mh{Although RLAIF has achieved impressive results in robot manipulation, the application of PbRL to} robot locomotion remains relatively unexplored because legged locomotion require optimizing multiple objectives simultaneously.
\mh{LAPP \cite{jian2025lapp} extends RLAIF to robot locomotion,} achieving faster convergence and improving overall performance. However, it still relies on \mh{the summation of trained rewards and} carefully engineered rewards to bootstrap and stabilize policy training. 

DAPPER \cite{kadokawa2026dapper} introduces a VLM-based framework to promote stable locomotion through automated trajectory evaluation and active learning to reduce annotation cost. Nevertheless, VLM-based feedbacks are still expensive to acquire and struggles to capture certain continuous control objectives, such as precise velocity tracking, particularly in robot systems where single-frame visual observations provide limited dynamic information.

In contrast, MAPL addresses these limitations by integrating multiple pretrained \mh{preference-scoring} models, each specializing in distinct locomotion aspects such as velocity tracking, stability, and motion smoothness. This structured multi-objective feedback formulation enables \mh{reward learning for subtle policy details} while reducing dependence on hand-crafted rewards.
\section{Preliminaries}
\label{sec:prelim}

\begin{algorithm}[t]

\caption{MAPL}
\label{alg:Pseudocode}
\begin{algorithmic}[1]
\Require Buffer $B$, Preference Scoring Model $\sigma_\psi$, Policy $\pi_\theta$, Scoring Model Update Frequency $T_{P}$

\State Populate $B$ with $L$ initial trajectories sampled from $\pi$

\For{$t = 1, 2, \dots$}
    \If{$t \% T_P == 0$} 
        \State Collect trajectories $\mathcal{D}$ sampled from $\pi$
        \State Generate preference rankings for each $\tau \in \mathcal{D}$
        \State Update scoring model parameters $\psi$ using Eq.~\ref{eq:loss}
    \EndIf
    \State Rollout policy for one step to obtain $s_t, a_{t-1}, s_{t-1}$
    \State Calculate reward, $r_{t} = \sigma_\psi(s_{t}) - \sigma_\psi(s_{t-1})$
    \State Update policy parameters $\theta$
\EndFor

\end{algorithmic}
\end{algorithm}

We consider the standard Markov Decision Process (MDP) reinforcement learning setting, where an agent sequentially interacts with an environment~\cite{sutton1998reinforcement}.
At each timestep $t$, the agent receives an observation $o_t$ from the current state $s_t$ and selects an action $a_t$ according to the policy $\pi$. 
After executing the action, the environment returns a reward $r_t$ and transitions to the next state $s_{t+1}$. 
The objective of the agent is to maximize the expected return, defined as the discounted sum of rewards:
\begin{equation}
R = \sum_{k=0}^{\infty} \gamma^k r(s_k, a_k),
\end{equation}
where $\gamma$ is the discount factor.

Our work builds upon the PbRL framework, where the agent learns a reward function from preference feedback instead of relying on handcrafted reward signals.
In traditional PbRL, a trajectory $\tau$ is defined as a sequence of states:
\begin{equation}
\tau = \{(s_t, a_t), (s_{t+1}, a_{t+1}), \ldots, (s_H, a_H)\}, \quad H \ge 1.
\end{equation}

During training, pairs of trajectories are sampled, and human annotators or AI models provide preference labels over trajectory pairs. The collected preference signals are then used to train a reward model that outputs scalar rewards, effectively replacing the environment reward.
In our method, we sample trajectory pairs $(\tau^0, \tau^1)$ during training, where each trajectory is downsampled so that it contains $M$ evenly-distributed states.
This allows us to capture long-horizon trends within a compact trajectory.
The preference label is defined as
\[
y =
\begin{cases}
0, & \tau^0 \succ \tau^1, \\
1, & \tau^1 \succ \tau^0, \\
0.5, & \tau^0 \sim \tau^1.
\end{cases}
\]

We follow the Bradley--Terry model \cite{bradley1952rank} to compute the likelihood of $\tau^1 \succ \tau^0$ for a pair of trajectories:
\begin{equation}
\label{eq:bt}
P_{\psi}(\tau^1 \succ \tau^0)
=
\frac{
\exp\left(\sum_{t=1}^{H} \sigma_{\psi}(s_t^1)\right)
}{
\sum_{i \in \{0,1\}}
\exp\left(\sum_{t=1}^{H} \sigma_{\psi}(s_t^i)\right)
}.
\end{equation}

We train a transformer-based reward model~\cite{kim2023preference} by minimizing the pairwise loss
\begin{equation}
\begin{aligned}
\label{eq:loss}
    \mathcal{L}^{(i)} &= -\mathbb{E}_{(\tau^0, \tau^1, y) \sim \mathcal{D}} \bigg[ \mathbb{I}\{y = (\tau^0 \succ \tau^1)\}\log P_\psi[\tau^0 \succ \tau^1] \\ &+ \mathbb{I}\{y = (\tau^1 \succ \tau^0)\} \log P_\psi[\tau^1 \succ \tau^0]\bigg].
\end{aligned}
\end{equation}

We denote the output as the \textit{preference score}, $\sigma_{\psi}(s_t)$.

\section{METHODOLOGY}
\label{sec:METHODOLOGY}

Figure~\ref{fig:overview} provides an overview of MAPL, which consists of two iterative steps, \textit{preference scoring model training} and \textit{policy training}, which operate at different frequencies.
The full MAPL framework is shown in Alg.~\ref{alg:Pseudocode}.

\subsection{Locomotion Objectives}

First, we consider a generic quadruped locomotion task that requires balancing multiple behavioral objectives under dynamic and contact-rich interactions. 
We identify three semantically grounded and task-agnostic criteria that are sufficient to characterize high-quality locomotion across diverse terrains: velocity tracking, stability, and motion smoothness.
We adopt the same three criteria across all terrain types without any task-specific modifications.
These criteria are defined entirely in natural language, and do not contain any domain-specific engineering.
Definitions are given below.

\textbf{Velocity Tracking.}
Measures how accurately the robot executes commanded linear and angular velocities.
\framebox{
\parbox{0.94\linewidth}{%
\textit{Prompt}: At each timestep, compare the robot’s actual x–y linear velocity to the commanded x–y velocity.
Smaller deviations correspond to better tracking, and preference should decrease smoothly as deviations increase.
Larger errors should reduce preference more strongly, but no single timestep should completely invalidate a trajectory.
}
}
\vspace{0.05em}

\textbf{Stability.}
Measures how well the robot is able to maintain an upright position, avoiding excessive roll and pitch.
\framebox{
\parbox{0.94\linewidth}{%
\textit{Prompt}: The base height should stay very close to \texttt{height} throughout the trajectory.
Even small deviations should noticeably reduce preference, and larger deviations should rapidly make the trajectory much worse, regardless of other factors.
The robot base should exhibit minimal vertical motion.
Roll and pitch angular velocities should remain small and smooth.
}
}
\vspace{0.05em}

\textbf{Smoothness.}
Measures the smoothness of control signals.
\framebox{
\parbox{0.94\linewidth}{%
\textit{Prompt}: We prefer the trajectory with smaller overall action difference.
Trajectories with fewer or weaker stumble events are strictly preferred over those with more frequent or more severe stumbles.
If one trajectory exhibits clear stumble behavior while the other does not, the non-stumbling trajectory should be preferred, even if its action smoothness is slightly worse.
}
}
\vspace{0.05em}

\subsection{Multi-Objective Preference Generation}

Existing RLAIF approaches \mh{collapse} multiple behavioral objectives into a single preference ranking.
However, this forces the LLM to reason over \mh{multiple} criteria simultaneously, leading to complex prompt engineering, \mh{noisy supervision signals, and gradient interference, especially when some objectives are more reliably estimated than others.}

Our work is based on the insight that LLMs elicit more accurate preference rankings if each objective is ranked and modeled independently.
To that end, we define a prompt in which the LLM produces a separate preference ranking for each objective defined above.
As specified in Sec.~\ref{sec:prelim}, the prompt is given a complete trajectory which has been downsampled to consist of $M$ evenly-distributed states.
To account for stochasticity and better capture the mode of the LLM sampling distribution, we query preference rankings multiple times and take the majority answer.

\subsection{\mh{Preference Scoring} Model Training}
In the preference scoring model training step, we train a transformer-based preference scoring model $\sigma$, parameterized by $\psi$, consisting of shared parameters and a separate output head for each objective, where the $i$-th head produces a preference score
\(
\sigma_\psi^{(i)}(s)
\), 
where \(i \in \{\text{velocity}, \text{stability}, \text{smoothness}\}\).
We adopt a pairwise ranking loss following the Bradley--Terry formulation in Equation \ref{eq:bt}.
The scoring model is updated at a slower frequency than the policy---in our experiments 100$\times$ slower---such that we avoid computationally expensive LLM calls until the policy's state visitation distribution has sufficiently evolved.

\begin{figure}[]
    \centering
    \includegraphics[width=\linewidth]{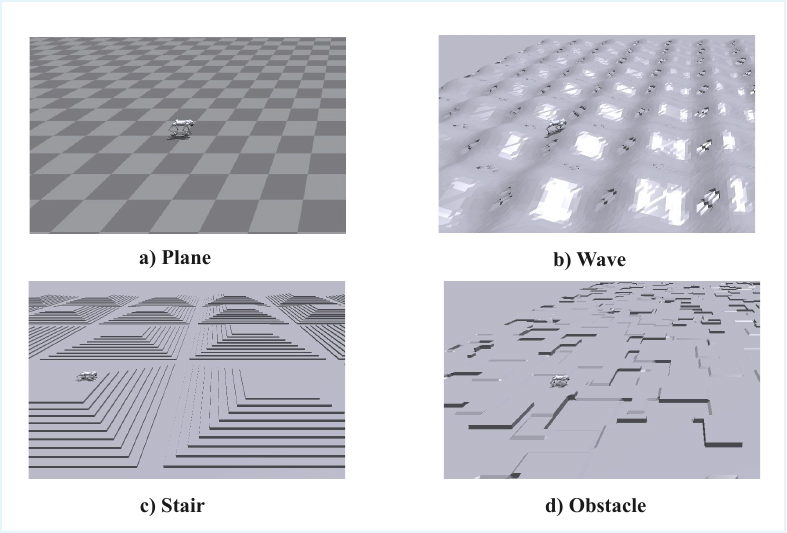}
    \caption{Evaluation terrains where the quadratic robots are trained to walk.}
    \label{fig:simulation environment}
\end{figure}

\subsection{Policy Training}
\label{sec:method-policy}

\begin{figure*}[]
\vskip -0.1in
    \centering
    \includegraphics[width=1\linewidth]{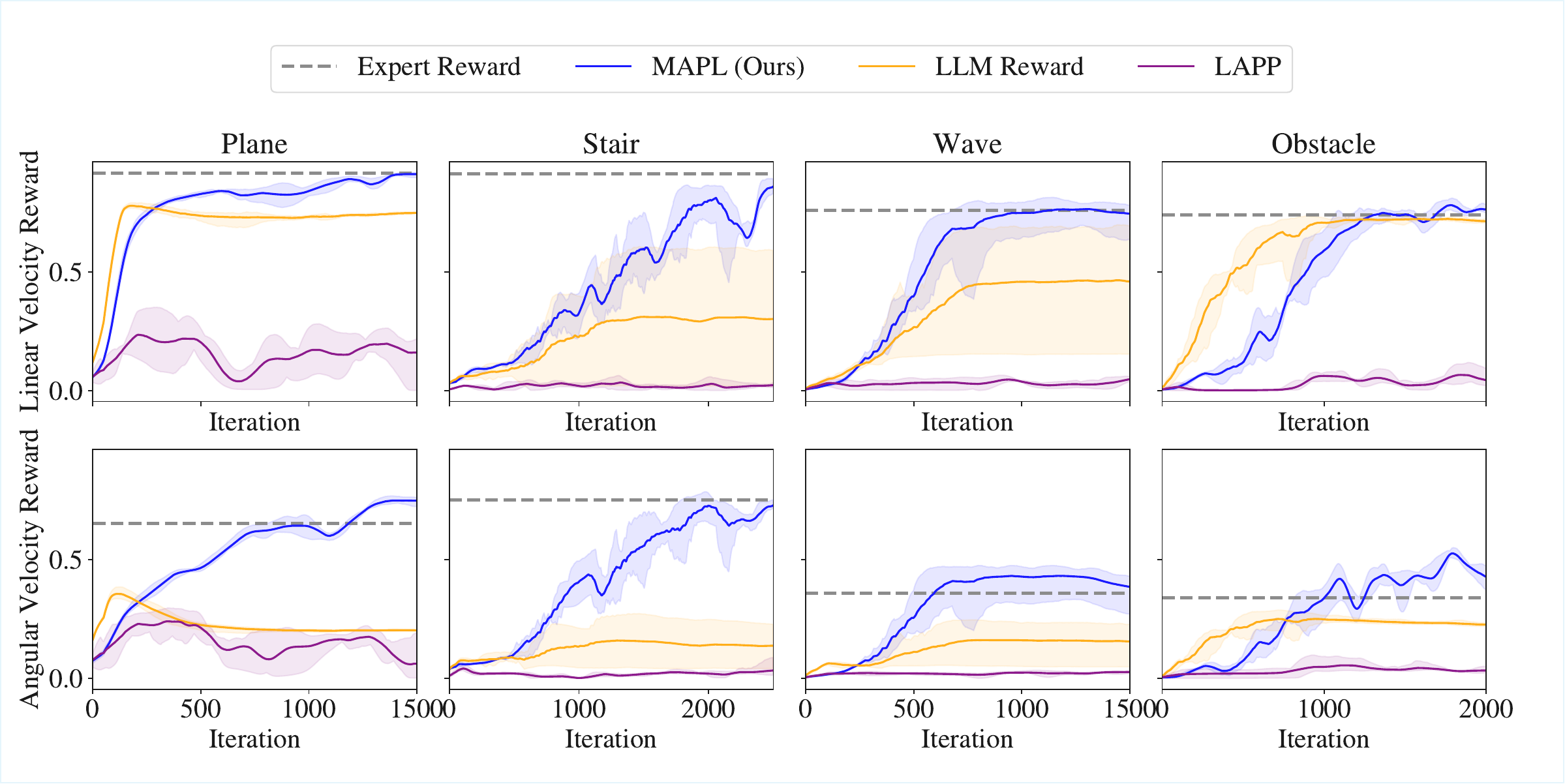}
    \caption{Velocity training curves of MAPL and baseline methods across diverse terrains. All results are averaged over five independent random seeds. The solid lines indicate the mean performance, while the shaded regions represent the standard error across runs. A moving average with a window size of 100 is applied to improve visualization clarity of the learning trends.}
    \label{fig:training_curves}
\end{figure*}

In the policy training step, state-action pairs are sampled by rolling out the policy in the environment.
Overall preference scores are computed for each state as a linear combination of per-objective preference scores,
\begin{equation}
    \sigma_\psi(s_t) = \sum_{i=1}^{N} \lambda_i \, \sigma_\psi^{(i)}(s_t),
\end{equation}
where $\lambda$ is a coefficient determining the relative weight of each objective.
This formulation allows us to control objective balancing through a low-dimensional weight vector rather than through fine-tuning a high-dimensional reward model.
This design decouples reward representation learning from objective balancing and improves training stability.
While $\lambda$ can be automatically computed by optimizing the policy's expected reward over a set of rollouts, in practice we treat it as an additional hyperparameter that is manually specified.

We prioritize velocity tracking and stability by assigning them large weights.
By contrast, smoothness is terrain-dependent.
In simple terrain types such as a flat plane, we assign a small weight value to avoid overly restrictive motion regularization.
In more challenging terrain such as stairs, we assign a larger weight to avoid undesirable events such as stumbling and abrupt contact transitions.

Rather than directly using the preference score as the reward signal for policy optimization, as in prior works~\cite{lin2024navigating, swamy2024minimaximalist} we adopt a potential difference formulation as the final reward to improve sample efficiency and policy return:
\begin{equation}
    r(s_t) = \sigma_\psi(s_t) - \sigma_\psi(s_{t-1}).
\end{equation}

\section{Experiments and Results}
\label{sec:experiment}
\subsection{Experiment Setup}

We evaluate MAPL by comparing downstream policy performance to baseline approaches, where the goal is to train a legged quadruped to run at a specified velocity across varied terrain.

\textbf{Environments. }
Our environments are implemented based on the Legged-Gym framework, which provides standardized terrain generation and locomotion task configurations~\cite{rudin2022learning}. Evaluation terrains include a flat plane, hills, pyramidal stairs, and randomly-sized obstacles.
We conduct all experiments on the Unitree Go2 quadruped robot using the Isaac Gym simulator. 
See Figure \ref{fig:simulation environment} for the visualization of these simulated environments.

\begin{figure*}[]
    \centering
    \includegraphics[width=0.95\linewidth]{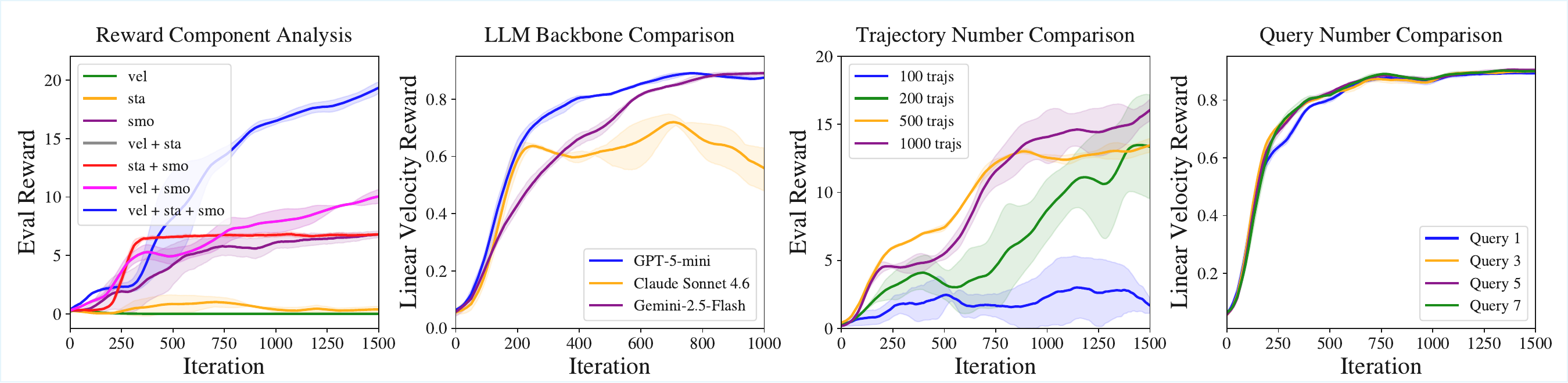}
    \caption{All results are averaged over three independent random seeds for the ablation studies. 
    Solid lines denote the mean training performance, and shaded regions indicate the standard error across runs. We conduct 4 abolation studies which includes: 1) Contribution analysis of each reward component in the MAPL framework to evaluated locomotion performance where vel, sta, and smo denote velocity, stability, and smoothness, respectively. 2) Comparison of different large language model backbones for velocity tracking within MAPL. 3) Sensitivity analysis with respect to the size of the trajectory replay buffer. 4) Sensitivity analysis with respect to the number of LLM queries per trajectory ranking. Eval reward represents expert reward values.}
    \label{fig:ablation}
\end{figure*} 

\begin{figure*}[]
    \centering
    \includegraphics[width=0.95\linewidth]{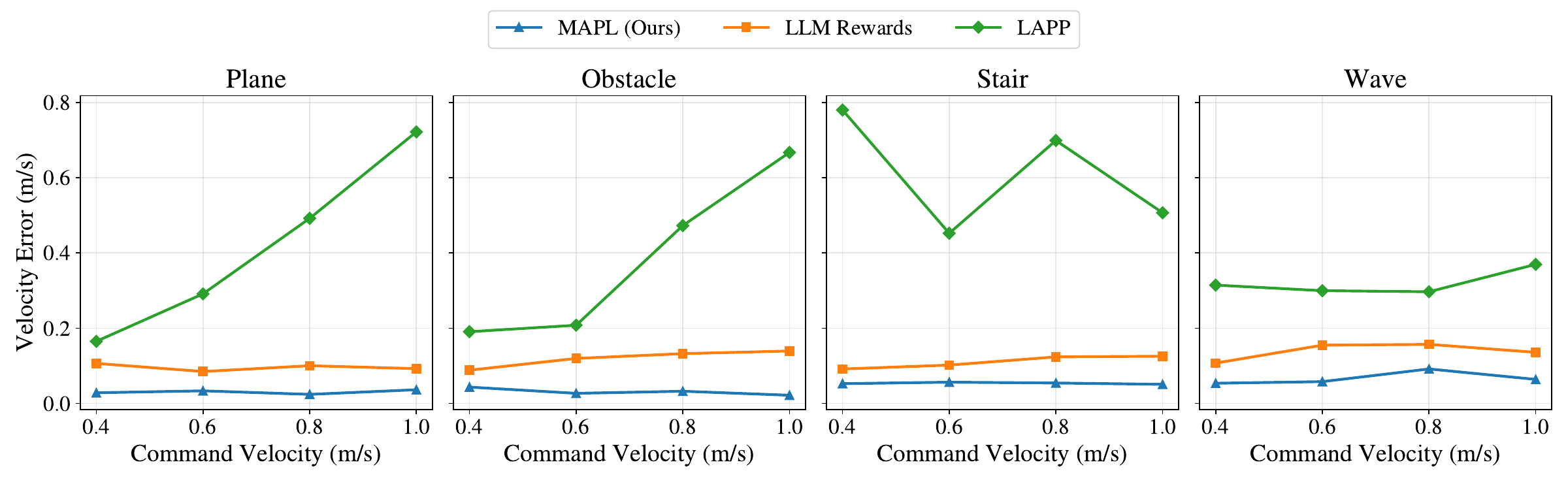}
    \caption{
        Velocity tracking error of the trained policies across different methods. The error is computed under given forward command velocities, where velocities are averaged over 20-step intervals during a 500-step rollout.
    }
    \label{fig:velocity_error}
\end{figure*}

\textbf{Metrics.}
We measure performance with two metrics:
\begin{itemize}
    \item \textbf{Linear Velocity Reward}.
    \begin{equation}
        r^{v_{xy}} =
        \exp\left(
        - \frac{
        \left\| \mathbf{v}_{xy}^{\text{cmd}} - \mathbf{v}_{xy}^{\text{base}} \right\|_2^2
        }{\sigma_v}
        \right)
    \end{equation}
    \item \textbf{Angular Velocity Reward}.
    \begin{equation}
        r^{v_\theta} =
        \exp\left(
        - \frac{
        \left( v_\theta^{\text{cmd}} - v_\theta^{\text{base}} \right)^2
        }{\sigma_\theta}
        \right)
    \end{equation}
    
    The parameters $\mathbf{v}^{\text{cmd}}$ denote the desired velocities, 
    $\mathbf{v}^{\text{base}}$ represent the measured base velocities of the quadruped robot; 
    $\sigma_v$ and $\sigma_\theta$ are scaling hyperparameters that control the tolerance to velocity tracking errors. 
    In all environments, both $\sigma$ are set up to be 0.25. Both rewards are Gaussian-shaped tracking terms encouraging accurate velocity control for locomotion. They issue higher values when the measured linear or angular velocity of the robot closely matches the commanded velocity, and exponentially penalize deviations.
\end{itemize}

\begin{table}[t]
\centering
\small
\begin{tabular}{p{11.19em} p{8.5em} p{2.15em}}
\hline
\textbf{Term} & \textbf{Equation} & \textbf{Weight} \\
\hline

$r_{v_{xy}}^{cmd}$: $v_{xy}$ tracking &
$e^{\left(-\frac{\|\mathbf{v}_{xy}-\mathbf{v}_{xy}^{cmd}\|^2}{\sigma_v}\right)}$ &
1.0 \\

$r_{\omega_z}^{cmd}$: yaw tracking &
$e^{\left(-\frac{(\omega_z-\omega_z^{cmd})^2}{\sigma_\omega}\right)}$ &
0.5 \\

$z$ velocity &
$v_z^2$ &
$-2.0$ \\

roll-pitch velocity &
$\|\boldsymbol{\omega}_{xy}\|^2$ &
$-5e^{-2}$ \\

joint torques &
$\|\tau\|^2$ &
$-1e^{-5}$ \\

joint accelerations &
$\|\ddot q\|^2$ &
$-2.5e^{-7}$ \\

thigh/calf collision &
$1_{\text{collision}}$ &
$-1.0$ \\

joint limit violation &
$1_{q_i>q_{max}\,||\,q_i<q_{min}}$ &
$-10$ \\

feet air time &
$\sum_{foot} t_{air}$ &
1.0 \\

action smoothing &
$\|a_{t-1}-a_t\|^2$ &
$-1e^{-2}$ \\

\hline
\end{tabular}
\caption{The hand-designed expert reward function consists of a linear combination of the above terms.}
\label{tab:reward_terms}
\end{table}

\textbf{Curriculum and Difficulty-Weighted Evaluation.}
For uneven terrains (e.g., stairs, waves, and obstacles), we adopt a curriculum learning strategy. The terrain difficulty is gradually increased only after the policy reaches a predefined success threshold, indicating stable locomotion under the current difficulty level.
To ensure fair comparison across baselines trained under varying terrain difficulties, we report a difficulty-weighted tracking reward during evaluation. Specifically, the velocity reward is scaled proportionally to the normalized difficulty level:
\begin{equation}
\tilde{r} = r \cdot \frac{d + 1}{D_{\max}},
\end{equation}
where $d$ denotes the current terrain difficulty level, $D_{\max}$ is the maximum difficulty level, and $r$ represents the raw velocity tracking reward.
This scaling ensures that policies capable of handling higher-difficulty terrains receive proportionally higher evaluation rewards. When the highest difficulty level is reached ($d = D_{\max} - 1$), the scaling factor becomes 1, and no additional weighting is applied.

\textbf{Baselines.}
We compare against the following baselines:

\begin{itemize}
    \item \textbf{LAPP}~\cite{jian2025lapp}. Representative preference learning method based on a single preference signal. It employs a transformer-based reward model and uses carefully designed prompts with in-context examples to guide the LLM toward producing accurate feedback.
    Unlike the original version, for fair comparison we use a variant which does \textit{not} incorporate the standard environment reward.
    \item \textbf{LLM Reward}. Representative code-based baseline, in which an LLM is prompted to generate code for a reward function given the same prompts as MAPL.
    \item \textbf{Expert Reward}. Hand-designed reward function created by a domain expert, defined in Table~\ref{tab:reward_terms}. Represents upper-bound performance in policy training.
\end{itemize}

All baselines use the PPO-based \cite{schulman2017proximal} RSL-RL as the underlying reinforcement learning algorithm, following the network architecture and hyperparameter settings reported in prior work \cite{schwarke2025rslrl}. For the preference aggregation of MAPL, we set 
\(\lambda_{\text{velocity}}, \lambda_{\text{stability}}, \lambda_{\text{smoothness}}\) ratio to be 1.5, 1.0, and 0.1 for flat and wave terrains, and 1.5, 1.0, 0.25 for stair and obstacle terrains.
Unless specified otherwise, all rankings are produced with GPT-5-mini.

\begin{figure*}[]
    \centering
    \includegraphics[width=0.95\linewidth]{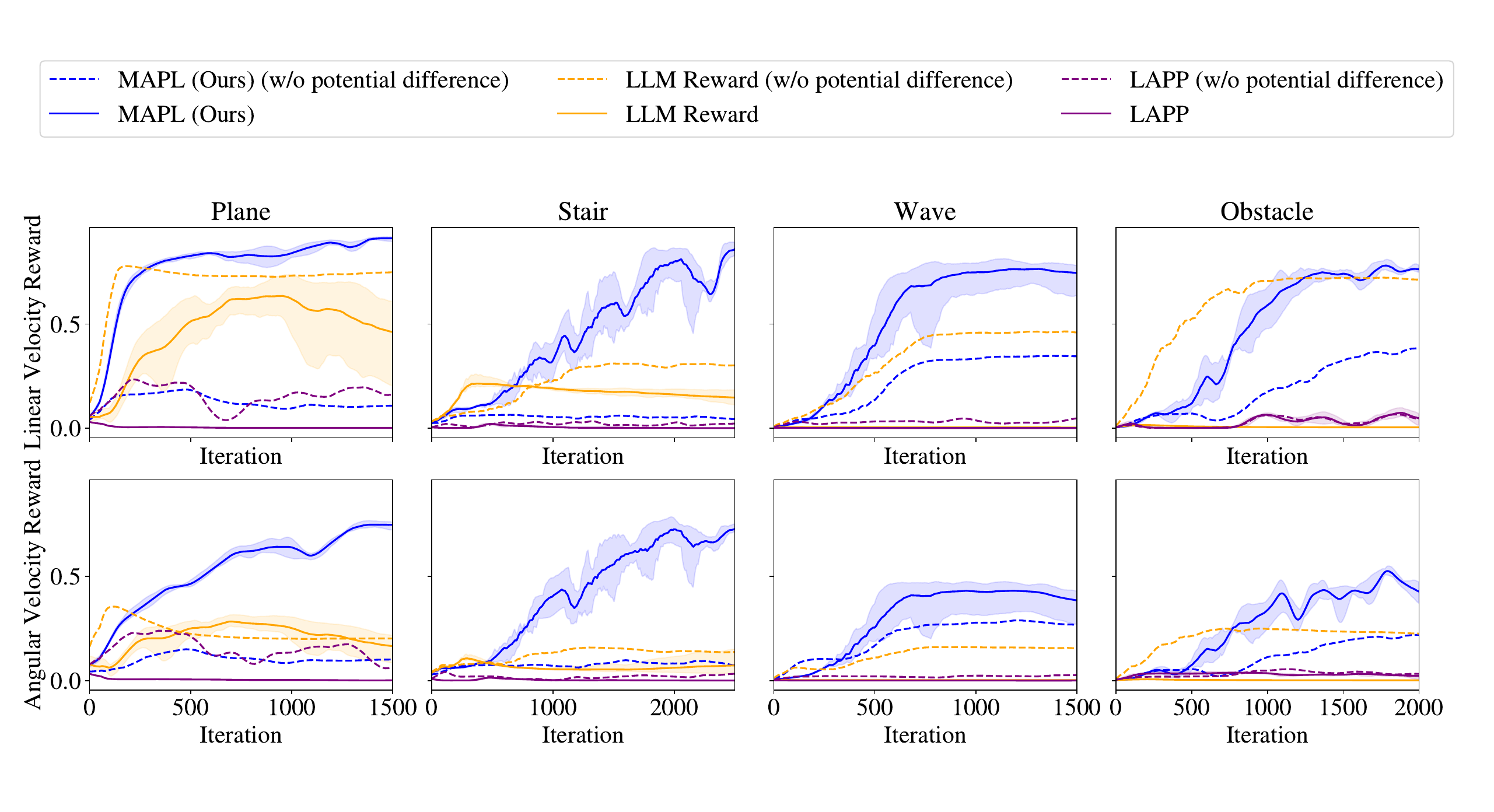}
\vskip -0.2in
    \caption{Velocity learning curves of all methods with and without the potential difference. The results are running for 5 seeds, and the moving average with window size 100 are applied in the figure to improve the readability.}
    \label{fig:pot_diff}
\end{figure*}

\subsection{Locomotion Policy Performance}
\label{sec:loco-result}
We next evaluate how the proposed reward modeling approach performs compared to baselines with GPT-5-mini in all environments. We collected 500 trajectory pairs for training each preference-based reward model. Each rewarding method is employed to train 5 RL policies with random seeds and initializations for each method. Training performance is measured by the residual error between commanded and actual velocities. Figure \ref{fig:training_curves} shows the resulting learning curves. The dashed gray line denotes the PPO baseline with well-tuned rewards, which is expected to be the upper-bound performance.

Across all terrains, our method (blue) consistently achieves higher final tracking rewards while LLM reward plateaus at lower returns, even if the policy training speed is limited by multi-criterion rewards and sometimes becomes lower than that of LLM Rewards. Without the assistance of well-tuned rewards, LAPP rewards alone fail in learning useful policies. Our trained reward models reach the converging policy returns of well-tune expert rewards in all scenarios and even outperform them over some terrains.

\subsection{Preference Component Ablation}
To understand the contribution and necessity of each preference component in the reward assembly, we conduct an ablation study. Keeping other settings unchanged, we train locomotion policies using rewards constructed from different subsets of the proposed preference terms, velocity, stability, and smoothness, as well as their combinations. Each reward model is trained using rankings over the same 2000 trajectory pairs to ensure convergence under diverse preference objectives. Figure \ref{fig:ablation} compares the learning curves over training iterations.

Without the robustness requirements, velocity preference alone fails to produce meaningful locomotion policies, whose performance remains near zero. The stability-only variant controls the legged robot slightly better, but fails to achieve obvious returns. The smoothness-only variant helps the policy capture more details but still converge at low returns.

Pairwise combinations significantly improve learning but remain inoptimal. Since the smoothness-only preference independently learns meaningful policies, it is reasonable to observe that combining it with other preference scores yields the greatest improvement in policy training compared to using any single preference alone. Compared to the returns with smoothness-only preference, adding velocity variant to measure the locomotion outcome increases training speed and returns simultaneously. Adding stability preference to the smoothness one boosts the learning speed significantly, but the conservative strategy leads to similar converging returns. Without the information from smoothness preference, velocity + stability fails in leading to any improvement.

The full reward achieves the best performance, nearly 2× higher than the best two-term variant with reduced variance. velocity drives forward motion, stability enforces physically robust body dynamics, and smoothness mitigates control oscillations. Removing any component disrupts this balance, degrading both convergence speed and returns. These results validate our multi-objective preference-based reward design.

\subsection{Trajectory and Query Sensitivity}
We further investigate the sensitivity of our method to the size of the trajectory buffer in the plane environment (top-left in Fig.~\ref{fig:ablation}) \mh{while keeping all other settings the same}. \mh{While MAPL rewards are used to train policies, "Eval Rewards" here used to measure policy training returns is the expert reward function.}
As shown in the figure, using 1000 trajectories yields the highest evaluation reward. 
When the trajectory number is reduced to 500, the performance remains competitive, \mh{implying the sample efficiency of MAPL. }
\mh{We also study the query number of one trajectory ranking required by MAPL with GPT-5-mini (bottom-right in Fig.~\ref{fig:ablation}) over the Plane terrain while keeping other settings the same. Due to hallucination, an LLM may produce inconsistent responses to repeated queries of the same question when it is uncertain, which constitutes a critical source of errors. A common mitigation strategy is to query the LLM multiple times and use majority voting as the final answer. By comparing training results under different numbers of queries per trajectory ranking, we evaluate the robustness of MAPL to such LLM-induced ranking errors.
The results show that the policy trained with a single query per ranking converges to nearly the same return within almost the same training steps as policies trained with multiple queries, although its early-stage learning is slightly slower. When the number of queries is greater than or equal to three, the learning curves become almost identical, where increasing the number of queries yields no noticeable performance improvement. These findings indicate that MAPL is robust to LLM inconsistent ranking answers and does not require a large number of queries per ranking.} 

\subsection{Potential Difference Reward Modeling}
To further boost the training speed and final returns of locomotion policies, MAPL has utilized the technique of potential difference. To demonstrate its effectiveness, Figure \ref{fig:pot_diff} compares training performance of all methods with and without potential-difference shaping across four locomotion tasks while keeping all other settings same as previous sections. Across all environments, removing the potential difference from MAPL consistently results in substantial degradation in both final performance and convergence speed. In contrast, incorporating potential difference enables MAPL to outperform all baselines, as detailed in Section~\ref{sec:loco-result}. Notably, integrating potential difference into LLM Reward or LAPP often leads to early plateauing or unstable training. This suggests that these methods fail to utilize potential-based shaping, likely due to poorly structured rewards induced by LLM hallucinations or single-criterion preference models.

\subsection{Robot Performance under Different Rewards}
\begin{figure}[]
\vspace{-3ex}
\vskip -0.2in
    \centering
    \includegraphics[width=\linewidth]{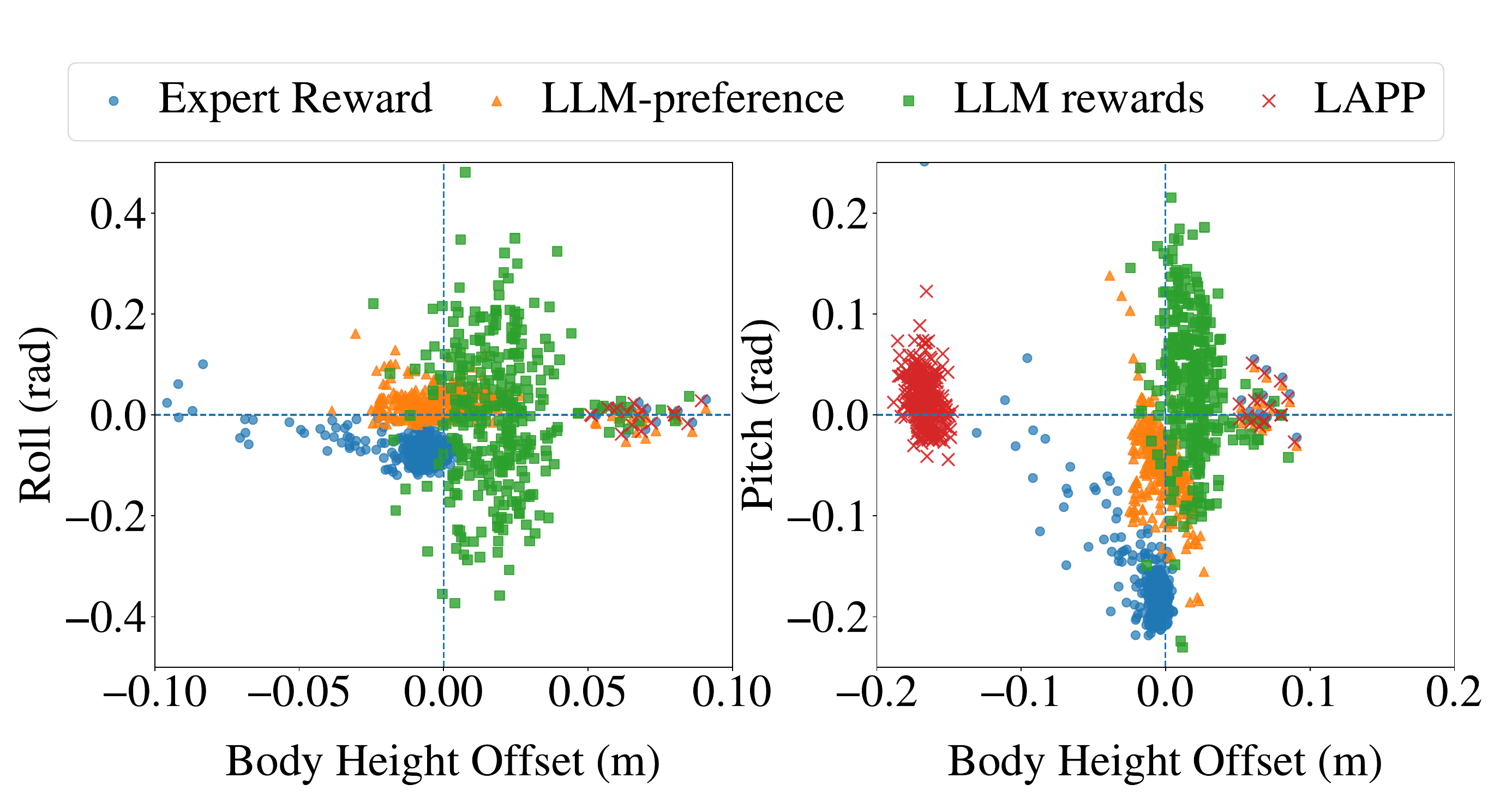}
    \caption{The robot height, roll, pitch distribution over 20 random steps rolled out by policies trained with different rewards in 16 randomly initialized Plane environment objectives.}
    \label{fig:height}
\end{figure}

To demonstrate the actual effects of different rewards over the robot, we compare the distributions of body height offset, roll, and pitch under different reward designs. In this experiment, 20 steps are randomly sampled by rolling out the policy trained with different rewards in 16 randomly-initialized Plane environment objectives respectively. We record the roll, pitch and robot body height at each step for each method, which are plotted in Figure \ref{fig:height}. Given that zero height offset and near-zero roll and pitch imply stable standing, points closer to the origin indicate better robot performance. As expected, expert rewards produce a relatively concentrated distribution around the origin. However, our reward achieves a tight clustering even closer to the origin, demonstrating improved stability even without expert-crafted terms and manual tuning. In contrast, LLM rewards without preference modeling exhibit higher variance due to LLM hallucination and its limited numerical concepts. Meanwhile, LAPP shows clear bias away from the origin, indicating unstable posture. This is consistent with its low training returns and ill-posed rewards without the assistance of well-tuned rewards, which are discussed in previous sections. Consistent with Section \ref{sec:loco-result}, these results illustrate how our method not only matches but surpasses expert-designed rewards, LLM generated rewards and single-preference rewards. They validate the effectiveness of structured preference aggregation and potential-difference shaping for properly rewarding robust and balanced locomotion behaviors.

\begin{figure}
\vspace{-2ex}
\vskip -0.2in
    \centering
    \includegraphics[width=1\linewidth]{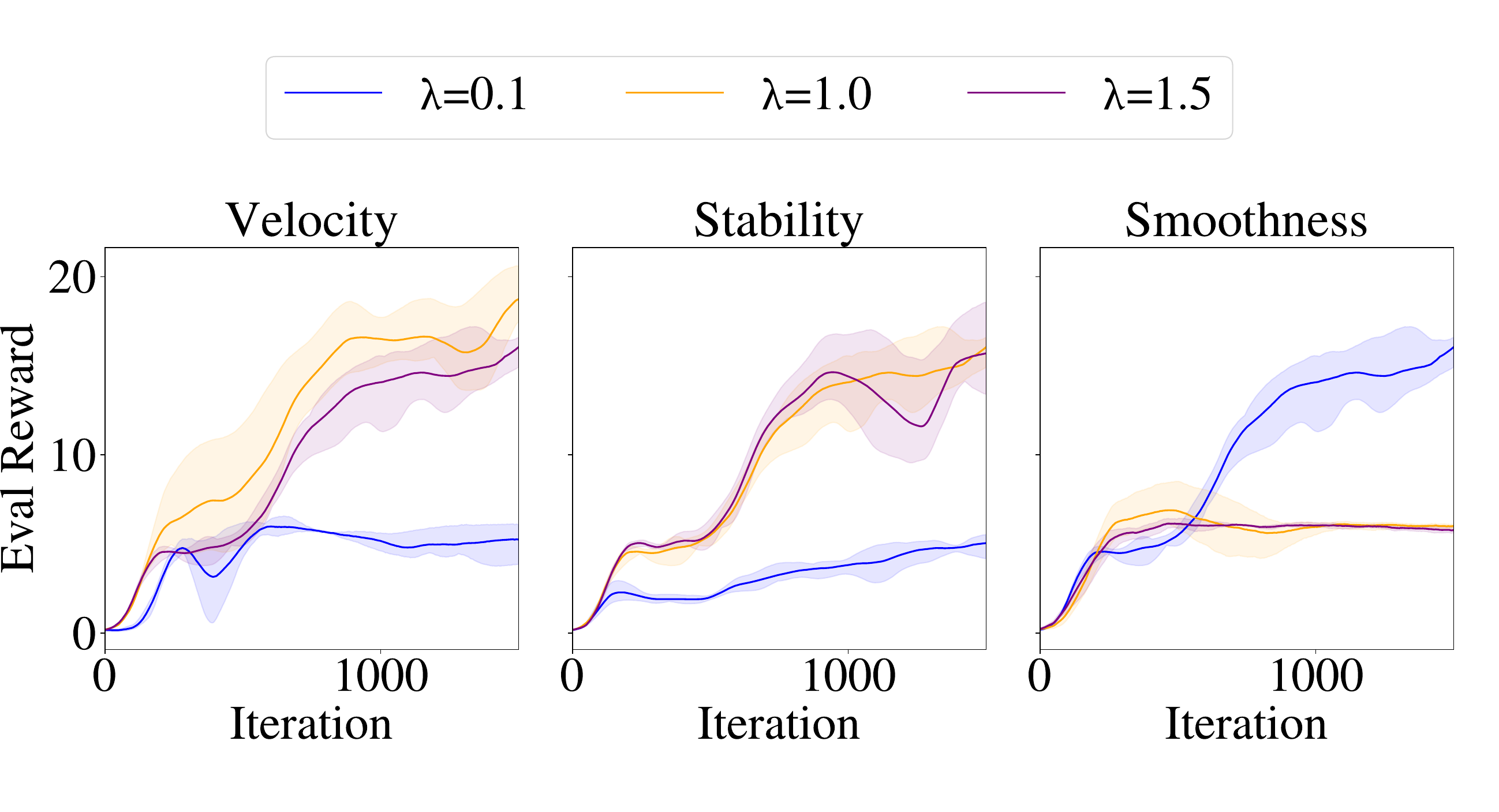}
\vspace{-2ex}
\vskip -0.1in
    \caption{Evaluation rewards under different $\lambda$ settings for each preference objective. Results are averaged over three random seeds with a smoothing window of 100. The shaded areas indicate the standard error.}
    \label{fig:lambda}
\end{figure}

\subsection{Preference Aggregation Weight Sensitivity}
We analyze the sensitivity of reward quality to the aggregation weight $\lambda$ for each preference criterion by comparing the resulting policy performance. As shown in Figure \ref{fig:lambda}, for velocity and stability, large weights with distinct scales lead to high but similar final returns, consistent with the dominant role of these two criteria in shaping the policy. On the other hand, consistent to the fact that smoothness plays a less important role, high $\lambda$ values achieve similarly low training returns. These results suggest that, the critical thing about $\lambda$ setting is identifying which preference criteria have stronger or weaker influence on the policy as discussed in Section \ref{sec:method-policy}, which is intuitive. As long as more influential objectives receive relatively larger weights, preference scores from multiple aspects can be combined reliably without extensive hyperparameter tuning.

\subsection{Velocity Error Analysis}
We evaluate velocity tracking performance by measuring the error between commanded and actual forward velocities under different command magnitudes. Specifically, we fix lateral and yaw commands to zero and vary the forward velocity from 0.4 to 1.0 m/s.

As shown in Fig. \ref{fig:velocity_error}, MAPL consistently achieves the lowest velocity error across all terrains, including plane, obstacle, stair, and wave environments. Notably, MAPL maintains low velocity error as the commanded velocity increases, indicating strong generalization and stable tracking performance even under more aggressive commands.

In contrast, LLM-based reward methods have moderate velocity errors, suggesting limited capability in accurately capturing velocity tracking objectives. LAPP shows significantly larger errors and a clear degradation trend, particularly on plane and obstacle terrains where the error grows rapidly as command velocity increases. This indicates that single-objective preference signals fail to provide sufficient supervision for precise control.

\subsection{Ranking Accuracy Analysis}

\begin{table}[t]
\centering
\begin{tabular}{lcccc}
\hline
 & Overall & Velocity & Stability & Smoothness \\
\hline
SOSQ & 63\% & $\times$ & $\times$ & $\times$ \\
MOSQ & $\times$ & 67\% & 67\% & 98\% \\
MOMQ (MAPL) & $\times$ & 88\% & 76\% & 98\% \\
LAPP & 68\% & $\times$ & $\times$ & $\times$ \\
\hline
\end{tabular}
\caption{
Ranking accuracy of different preference-query strategies. SOSQ denotes a single prompt that combines all preference criteria and produces one overall ranking. MOSQ denotes a single prompt that outputs separate rankings for different objectives. MOMQ (MAPL) uses multiple objective-specific prompts and multiple queries to obtain separate rankings. LAPP uses a manually designed prompt to generate a single overall ranking.
}
\label{tab:ranking}
\end{table}
To demonstrate the strength of aggregating multi-objective preference scores from trained scoring models, we analyze the ranking accuracy of different preference-query strategies by comparing their predicted rankings with the ground-truth rewards by rolling out 100 trajectory pairs. For the single-objective strategies, the ranking is evaluated using the full expert reward function as the reference. For the multi-objective strategies, the complete expert reward function is decomposed into three components corresponding to velocity, stability, and smoothness, and the ranking accuracy is evaluated separately for each objective.

As shown in Table~\ref{tab:ranking}, our method (MOMQ), which employs multiple objective-specific prompts and multiple queries, achieves the most stable and accurate rankings across different objectives compared with the other strategies.
\section{CONCLUSIONS}
\label{sec:conclusi}

In this work, we introduced MAPL, a multi-objective AI-informed preference learning framework that learns locomotion rewards aggregating multi-criterion LLM feedback rather than manually engineered reward functions. By querying an LLM to rank trajectories along velocity tracking, smoothness, and stability, MAPL learns objective-wise preference models and then combines them linearly under a potential-difference formulation to provide a stable training signal. Across four locomotion environments, including flat ground, uneven terrain, stairs, and obstacle traversal, MAPL trains policies using only LLM preference supervision and achieves performance comparable to or better than expert-designed rewards. Importantly, a single generic prompt is sufficient to train policies across all terrains, highlighting MAPL ability to reduce the need for reward engineering. Meanwhile, the reward-component ablation study and LLM trajectory ranking accuracy analysis support the necessity of multi-objective preference modeling.

Despite these significant results, several limitations remain. The current framework aggregates objectives using a linear combination, which improves interpretability but may limit expressiveness. Future work could explore nonlinear aggregation mechanisms that better capture interactions among locomotion objectives. 
Finally, while LLM-preference-based reward models provide scalable supervision, integrating multimodal foundation models, such as vision-language and tactile-aware models, could further enable reward learning from visual perception and physical interaction signals. These directions could potentially extend the approach to more complex robot behaviors such as loco-manipulation.
\bibliographystyle{IEEEtran}
\bibliography{reference}

@article{christiano2017deep,
  title={Deep reinforcement learning from human preferences},
  author={Christiano, Paul F and Leike, Jan and Brown, Tom and Martic, Miljan and Legg, Shane and Amodei, Dario},
  journal={Advances in neural information processing systems},
  volume={30},
  year={2017}
}

@inproceedings{lin2024navigating,
  title={Navigating Noisy Feedback: Enhancing Reinforcement Learning with Error-Prone Language Models},
  author={Lin, Muhan and Shi, Shuyang and Guo, Yue and Chalaki, Behdad and Tadiparthi, Vaishnav and Pari, Ehsan Moradi and Stepputtis, Simon and Campbell, Joseph and Sycara, Katia},
  booktitle={Findings of the Association for Computational Linguistics: EMNLP 2024},
  year={2024}
}

@inproceedings{margolis2023walk,
  title={Walk these ways: Tuning robot control for generalization with multiplicity of behavior},
  author={Margolis, Gabriel B and Agrawal, Pulkit},
  booktitle={Conference on Robot Learning},
  pages={22--31},
  year={2023},
  organization={PMLR}
}

@article{zhou2026learning,
  title={Learning Tactile-Aware Quadrupedal Loco-Manipulation Policies},
  author={Zhou, Pokuang and Zhou, Yuhao and Luu, Quan Khanh and Han, Seungho and Zhang, Heng and Huang, Binghao and Li, Yunzhu and Ajoudani, Arash and Xu, Zhengtong and She, Yu},
  journal={arXiv preprint arXiv:2604.27224},
  year={2026}
}

@article{lee2020learning,
  title={Learning quadrupedal locomotion over challenging terrain},
  author={Lee, Joonho and Hwangbo, Jemin and Wellhausen, Lorenz and Koltun, Vladlen and Hutter, Marco},
  journal={Science robotics},
  volume={5},
  number={47},
  pages={eabc5986},
  year={2020},
  publisher={American Association for the Advancement of Science}
}

@article{margolis2024rapid,
  title={Rapid locomotion via reinforcement learning},
  author={Margolis, Gabriel B and Yang, Ge and Paigwar, Kartik and Chen, Tao and Agrawal, Pulkit},
  journal={The International Journal of Robotics Research},
  volume={43},
  number={4},
  pages={572--587},
  year={2024},
  publisher={SAGE Publications Sage UK: London, England}
}

@article{kumar2021rma,
  title={Rma: Rapid motor adaptation for legged robots},
  author={Kumar, Ashish and Fu, Zipeng and Pathak, Deepak and Malik, Jitendra},
  journal={arXiv preprint arXiv:2107.04034},
  year={2021}
}

@article{park2022surf,
  title={Surf: Semi-supervised reward learning with data augmentation for feedback-efficient preference-based reinforcement learning},
  author={Park, Jongjin and Seo, Younggyo and Shin, Jinwoo and Lee, Honglak and Abbeel, Pieter and Lee, Kimin},
  journal={arXiv preprint arXiv:2203.10050},
  year={2022}
}

@article{lee2023rlaif,
  title={Rlaif vs. rlhf: Scaling reinforcement learning from human feedback with ai feedback},
  author={Lee, Harrison and Phatale, Samrat and Mansoor, Hassan and Mesnard, Thomas and Ferret, Johan and Lu, Kellie and Bishop, Colton and Hall, Ethan and Carbune, Victor and Rastogi, Abhinav and others},
  journal={arXiv preprint arXiv:2309.00267},
  year={2023}
}

@article{bai2022constitutional,
  title={Constitutional ai: Harmlessness from ai feedback},
  author={Bai, Yuntao and Kadavath, Saurav and Kundu, Sandipan and Askell, Amanda and Kernion, Jackson and Jones, Andy and Chen, Anna and Goldie, Anna and Mirhoseini, Azalia and McKinnon, Cameron and others},
  journal={arXiv preprint arXiv:2212.08073},
  year={2022}
}

@article{jian2025lapp,
  title={LAPP: Large Language Model Feedback for Preference-Driven Reinforcement Learning},
  author={Jian, Pingcheng and Wei, Xiao and Liu, Yanbaihui and Moore, Samuel A and Zavlanos, Michael M and Chen, Boyuan},
  journal={arXiv preprint arXiv:2504.15472},
  year={2025}
}

@article{wang2025primt,
  title={PRIMT: Preference-based Reinforcement Learning with Multimodal Feedback and Trajectory Synthesis from Foundation Models},
  author={Wang, Ruiqi and Zhao, Dezhong and Yuan, Ziqin and Shao, Tianyu and Chen, Guohua and Kao, Dominic and Hong, Sungeun and Min, Byung-Cheol},
  journal={arXiv preprint arXiv:2509.15607},
  year={2025}
}

@article{wang2024rl,
  title={Rl-vlm-f: Reinforcement learning from vision language foundation model feedback},
  author={Wang, Yufei and Sun, Zhanyi and Zhang, Jesse and Xian, Zhou and Biyik, Erdem and Held, David and Erickson, Zackory},
  journal={arXiv preprint arXiv:2402.03681},
  year={2024}
}

@inproceedings{wang2022skill,
  title={Skill preferences: Learning to extract and execute robotic skills from human feedback},
  author={Wang, Xiaofei and Lee, Kimin and Hakhamaneshi, Kourosh and Abbeel, Pieter and Laskin, Michael},
  booktitle={Conference on robot learning},
  pages={1259--1268},
  year={2022},
  organization={PMLR}
}

@article{lee2021pebble,
  title={Pebble: Feedback-efficient interactive reinforcement learning via relabeling experience and unsupervised pre-training},
  author={Lee, Kimin and Smith, Laura and Abbeel, Pieter},
  journal={arXiv preprint arXiv:2106.05091},
  year={2021}
}

@article{skalse2022defining,
  title={Defining and characterizing reward gaming},
  author={Skalse, Joar and Howe, Nikolaus and Krasheninnikov, Dmitrii and Krueger, David},
  journal={Advances in Neural Information Processing Systems},
  volume={35},
  pages={9460--9471},
  year={2022}
}

@article{kadokawa2026dapper,
  title={DAPPER: Discriminability-Aware Policy-to-Policy Preference-Based Reinforcement Learning for Query-Efficient Robot Skill Acquisition},
  author={Kadokawa, Yuki and Frey, Jonas and Miki, Takahiro and Matsubara, Takamitsu and Hutter, Marco},
  journal={IEEE Robotics \& Automation Magazine},
  year={2026},
  publisher={IEEE}
}

@article{ma2023eureka,
  title={Eureka: Human-level reward design via coding large language models},
  author={Ma, Yecheng Jason and Liang, William and Wang, Guanzhi and Huang, De-An and Bastani, Osbert and Jayaraman, Dinesh and Zhu, Yuke and Fan, Linxi and Anandkumar, Anima},
  journal={arXiv preprint arXiv:2310.12931},
  year={2023}
}

@article{zeng2024learning,
  title={Learning reward for robot skills using large language models via self-alignment},
  author={Zeng, Yuwei and Mu, Yao and Shao, Lin},
  journal={arXiv preprint arXiv:2405.07162},
  year={2024}
}

@article{heng2025boosting,
  title={Boosting universal llm reward design through heuristic reward observation space evolution},
  author={Heng, Zen Kit and Zhao, Zimeng and Wu, Tianhao and Wang, Yuanfei and Wu, Mingdong and Wang, Yangang and Dong, Hao},
  journal={arXiv preprint arXiv:2504.07596},
  year={2025}
}

@article{tu2024online,
  title={Online preference-based reinforcement learning with self-augmented feedback from large language model},
  author={Tu, Songjun and Sun, Jingbo and Zhang, Qichao and Lan, Xiangyuan and Zhao, Dongbin},
  journal={arXiv preprint arXiv:2412.16878},
  year={2024}
}

@article{wang2025prefclm,
  title={Prefclm: Enhancing preference-based reinforcement learning with crowdsourced large language models},
  author={Wang, Ruiqi and Zhao, Dezhong and Yuan, Ziqin and Obi, Ike and Min, Byung-Cheol},
  journal={IEEE Robotics and Automation Letters},
  volume={10},
  number={3},
  pages={2486--2493},
  year={2025},
  publisher={IEEE}
}

@inproceedings{venkataraman2025real,
  title={Real-world offline reinforcement learning from vision language model feedback},
  author={Venkataraman, Sreyas and Wang, Yufei and Wang, Ziyu and Ravie, Navin Sriram and Erickson, Zackory and Held, David},
  booktitle={2025 IEEE/RSJ International Conference on Intelligent Robots and Systems (IROS)},
  pages={13452--13459},
  year={2025},
  organization={IEEE}
}

@inproceedings{rudin2022learning,
  title={Learning to walk in minutes using massively parallel deep reinforcement learning},
  author={Rudin, Nikita and Hoeller, David and Reist, Philipp and Hutter, Marco},
  booktitle={Conference on robot learning},
  pages={91--100},
  year={2022},
  organization={PMLR}
}

@article{schwarke2025rslrl,
  title={RSL-RL: A Learning Library for Robotics Research},
  author={Schwarke, Clemens and Mittal, Mayank and Rudin, Nikita and Hoeller, David and Hutter, Marco},
  journal={arXiv preprint arXiv:2509.10771},
  year={2025}
}

@article{kim2023preference,
  title={Preference transformer: Modeling human preferences using transformers for rl},
  author={Kim, Changyeon and Park, Jongjin and Shin, Jinwoo and Lee, Honglak and Abbeel, Pieter and Lee, Kimin},
  journal={arXiv preprint arXiv:2303.00957},
  year={2023}
}

@article{schulman2017proximal,
  title={Proximal policy optimization algorithms},
  author={Schulman, John and Wolski, Filip and Dhariwal, Prafulla and Radford, Alec and Klimov, Oleg},
  journal={arXiv preprint arXiv:1707.06347},
  year={2017}
}

@article{bradley1952rank,
  title={Rank analysis of incomplete block designs: I. the method of paired comparisons},
  author={Bradley, Ralph Allan and Terry, Milton E},
  journal={Biometrika},
  volume={39},
  number={3/4},
  pages={324--345},
  year={1952},
  publisher={JSTOR}
}

@book{sutton1998reinforcement,
  title={Reinforcement learning: An introduction},
  author={Sutton, Richard S and Barto, Andrew G and others},
  volume={1},
  number={1},
  year={1998},
  publisher={MIT press Cambridge}
}

@article{wilson2012bayesian,
  title={A bayesian approach for policy learning from trajectory preference queries},
  author={Wilson, Aaron and Fern, Alan and Tadepalli, Prasad},
  journal={Advances in neural information processing systems},
  volume={25},
  year={2012}
}

@inproceedings{yu2025rlaif,
  title={Rlaif-v: Open-source ai feedback leads to super gpt-4v trustworthiness},
  author={Yu, Tianyu and Zhang, Haoye and Li, Qiming and Xu, Qixin and Yao, Yuan and Chen, Da and Lu, Xiaoman and Cui, Ganqu and Dang, Yunkai and He, Taiwen and others},
  booktitle={Proceedings of the Computer Vision and Pattern Recognition Conference},
  pages={19985--19995},
  year={2025}
}

@article{li2023silkie,
  title={Silkie: Preference distillation for large visual language models},
  author={Li, Lei and Xie, Zhihui and Li, Mukai and Chen, Shunian and Wang, Peiyi and Chen, Liang and Yang, Yazheng and Wang, Benyou and Kong, Lingpeng},
  journal={arXiv preprint arXiv:2312.10665},
  year={2023}
}

@article{swamy2024minimaximalist,
  title={A minimaximalist approach to reinforcement learning from human feedback},
  author={Swamy, Gokul and Dann, Christoph and Kidambi, Rahul and Wu, Zhiwei Steven and Agarwal, Alekh},
  journal={arXiv preprint arXiv:2401.04056},
  year={2024}
}
\newpage
\clearpage
\onecolumn

\appendix
\subsection{Reward Model Hyperparameter}
The hyparameter of Transformer-based reward model are tuned manually. The details are listed below.

\begin{table}[h]
\centering
\small
\begin{tabular}{l c}
\toprule
\textbf{Hyperparameter} & \textbf{Transformer Reward Model} \\
\midrule
Embedding Dimension & 256 \\
Number of Layers & 2 \\
Number of Heads & 4 \\
Feedforward Dimension & 512 \\
Dropout & 0.1 \\
Batch Size & 32 \\
Epochs & 80 \\
Learning Rate Schedule & 
\begin{tabular}[c]{@{}c@{}}
$[4\times10^{-5}, 20]$, \\
$[3\times10^{-5}, 70]$, \\
$[8\times10^{-5}, 80]$
\end{tabular} \\
\bottomrule
\end{tabular}
\caption{Training hyperparameters of the Transformer-based reward model. 
The learning rate schedule is presented as the learning rate value along with the corresponding epoch it is applied to.}
\label{tab:transformer_reward_hyper}
\end{table}

We adopt the same PPO hyperparameter to Legged gym \cite{rudin2022learning}:
\begin{table}[h]
\centering
\begin{tabular}{lc}
\toprule
\textbf{Hyperparameter} & \textbf{Value} \\
\midrule
Batch size & 98304 (4096 $\times$ 24) \\
Mini-batch size & 24576 (4096 $\times$ 6) \\
Number of epochs & 5 \\
Clip range & 0.2 \\
Entropy coefficient & 0.01 \\
Discount factor $\gamma$ & 0.99 \\
GAE discount factor $\lambda$ & 0.95 \\
Desired KL-divergence $kl^*$ & 0.01 \\
Learning rate $\alpha$ & adaptive$^{*}$ \\
\bottomrule
\end{tabular}
\caption{Training hyperparameters for PPO for all of our environment, ($^{*}$) means we use the adaptive learning rate based on the KL-divergence}
\end{table}

\subsection{System Prompt for Reward Type}
\label{sec:Appendix}

\subsubsection{Velocity Preference Prompt}

\setlength{\parskip}{0pt}
\setlength{\parindent}{0pt}
\setlength{\parskip}{0pt}
\setlength{\parindent}{0pt}
\begin{mdframed}[
    linewidth=0.8pt,
    roundcorner=4pt,
    innerleftmargin=8pt,
    innerrightmargin=8pt,
    innertopmargin=6pt,
    innerbottommargin=6pt
]

You are a robotics engineer specializing in analyzing and comparing the trajectories of a Unitree Go2 Robot Dog.
You will be given two trajectories of the Unitree Go2 Robot Dog, and you need to decide which trajectory is better.

A trajectory will include the following information:

1) "The base linear velocity" (m/s): Current Unitree Go2 robot dog x and y velocity, you will need that information to decide if the robot dog is moving forward, the shape of this term is (6, 2), where 6 is the time length, and 2 is x and y linear velocity respectively

2) "The base angular velocity" (rad/s): Current Unitree Go2 robot dog yaw velocity, you will need that information to decide if the robot dog is turning, the shape of this term is (6, 1)

3) "The commands"(m/s, m/s, rad/s): The desired x, y, yaw velocity of Unitree Go2 robot dog, you will need that information to decide if the robot dog is following the commands, the shape of this term is (6, 3)

4) "The feet contacts"[front left, front right, rear left, rear right]: The contact boolean values of the four feet on the ground. 1 means touching the ground while 0 means in the air. You only need that information for considering Gait pattern consistency. The shape of this term is (6, 4)

5) The negative sign only represents direction of velocity
 
\vspace{0.3em}
\textbf{Decision Rules}

1)Linear velocity tracking (primary criterion)
At each timestep, compare the robot’s actual x–y linear velocity to the commanded x–y velocity.
Smaller deviations correspond to better tracking, and preference should decrease smoothly as deviations increase.
Very small errors should still be slightly preferred over perfect tracking, but the difference should be minimal.
Larger errors should reduce preference more strongly, but no single timestep should completely invalidate a trajectory by itself.
Trajectories that maintain consistently low tracking error across timesteps are preferred over trajectories with higher average error.

2) Yaw (angular) velocity tracking (secondary criterion)
Evaluate how closely the robot’s yaw rate follows the commanded yaw rate across timesteps in the same smooth and continuous manner.

3) Gait pattern consistency (conditional criterion)
The robot is encouraged to use a diagonal gait (trot):
Front-left and rear-right feet tend to be in the same contact state at the same time.
Front-right and rear-left feet tend to be in the same contact state at the same time.

\vspace{0.3em}
\textbf{Overall preference rule}

The final decision must consider both linear and angular velocity tracking, but linear tracking should weight 3 times than angular tracking.
Gait or foot-contact patterns should only be considered when both linear and angular velocity tracking are already very close to the commanded values across timesteps.
If one trajectory shows clearly better linear or angular tracking, do not use gait differences to overturn that decision.

\vspace{0.3em}
\textbf{Output Format}

You will be given 20 pairs of trajectories each time, for each pair of trajectory

1) If the first trajectory is better than the second trajectory, the preference value for this pair of trajectory is 0.

2) If the second trajectory is better than the first trajectory, the preference value for this pair of trajectory is 1.

3) If the two trajectories are equally preferable, the preference value for this pair of trajectory is 2.

4) You should analyze each pair of trajectory independently, do not refer to previous results.
 
Please return a Python list of preference values. You must output ONLY a JSON array, do not output anything else, do not hallucinate or make up data, strictly follow the decision rules.

\end{mdframed}

\subsubsection{Stability Preference Prompt}

\begin{mdframed}[
    linewidth=0.8pt,
    roundcorner=4pt,
    innerleftmargin=8pt,
    innerrightmargin=8pt,
    innertopmargin=6pt,
    innerbottommargin=6pt
]

You are a robotics engineer specializing in analyzing and comparing the trajectories of a Unitree Go2 Robot.
You will be given two trajectories of the Unitree Go2 Robot, each trajectory consisting of continuous 6 states, and you need to decide which trajectory is better given 6 states in the trajectory.

A trajectory will include the following information (for STABILITY evaluation only):

1) "The base height" (m): The z position (height) of the robot base torso. Shape (6,1), one value per step. Height should stay close to 0.34 m with minimal fluctuation.

2) "The vertical linear velocity" (m/s): The z-axis component of the base linear velocity. Shape (6, 1). Values should be near 0 to avoid bouncing.

3) "The roll/pitch angular velocity" (rad/s): The base angular rates around x (roll) and y (pitch). Shape (6, 2). Magnitudes should be close to 0 to avoid rocking.

\vspace{0.3em}
\textbf{Decision Rules}

1)Base height consistency
The base height should stay very close to 0.34 m throughout the trajectory.
Even small deviations should noticeably reduce preference, and larger deviations should rapidly make the trajectory much worse, regardless of other factors.

2)Vertical motion suppression
The robot base should exhibit minimal vertical motion.
Occasional tiny vertical velocities are acceptable, but larger vertical oscillations should be penalized increasingly strongly, regardless of direction.

3)Roll / pitch motion smoothness
Roll and pitch angular velocities should remain small and smooth.
Persistent rocking or oscillation should gradually reduce preference, even if no single timestep is extreme.

\vspace{0.3em}
\textbf{Overall preference rule}

1)All three factors contribute jointly to stability quality

2)Large violations in any single factor should dominate the decision

3)Preference should degrade smoothly and continuously, not via hard thresholds.

\vspace{0.3em}
\textbf{Output Format}

You will be given 20 pairs of trajectories to compare each time, and you need to provide your preference value for each pair.

0) If the trajectory 0 is better, the preference value should be 0;

1) If the trajectory 1 is better, the preference value should be 1;

2) If the two trajectories are equally preferable, the preference value for this pair of trajectory is 2.

3) You should analyze each pair of trajectory independently, do not refer to previous results

Please return a Python list of preference values. You must output ONLY a JSON array, do not output anything else, do not hallucinate or make up data, strictly follow the decision rules.

\end{mdframed}

\subsubsection{Smoothness Preference Prompt}

\begin{mdframed}[
    linewidth=0.8pt,
    roundcorner=4pt,
    innerleftmargin=8pt,
    innerrightmargin=8pt,
    innertopmargin=6pt,
    innerbottommargin=6pt
]

You are a robotics engineer specializing in analyzing and comparing the trajectories of a Unitree Go2 Robot.
You will be given two trajectories of the Unitree Go2 Robot, each trajectory consisting of continuous 6 states, and you need to decide which trajectory is better given these states in the trajectory.

A trajectory will include the following information (for SMOOTHNESS evaluation only):

1) "sum of joint action  change $\Delta u$" (unitless): The total accumulated change in joint action commands across the entire trajectory. This value is already computed as the sum of per-step, per-joint action differences between consecutive action commands. The shape for this term is (6, 1)

2) "stumble": The number of foot slipping, scraping, or skidding, where the foot is not properly supporting the robot’s weight but is experiencing strong lateral forces. The shape for this term is (6, 1)

\vspace{0.3em}
\textbf{Decision Rules}

1)Action Smoothness: We prefer the trajectory with smaller overall joint action change.

2)Stumble avoidance: Trajectories with fewer or weaker stumble events are strictly preferred over those with more frequent or more severe stumbles.
If one trajectory exhibits clear stumble behavior while the other does not, the non-stumbling trajectory should be preferred, even if its action smoothness is slightly worse.

\vspace{0.3em}
\textbf{Overall preference rule}

1)Action smoothness and stumble behavior jointly determine trajectory quality.

2)Severe stumble behavior should dominate the decision.

3)When stumble differences are minor, action smoothness should be the primary deciding factor.

4)Preference should degrade smoothly and continuously, not via hard thresholds.

\vspace{0.3em}
\textbf{Output Format}

You will be given 20 pairs of trajectories to compare each time, and you need to provide your preference value for each pair.

0) If the trajectory 0 is better, the preference value should be 0;

1) If the trajectory 1 is better, the preference value should be 1;

2) If the two trajectories are equally preferable, the preference value for this pair of trajectory is 2.

3) You should analyze each pair of trajectory independently, do not refer to previous results

Please return a Python list of preference values. You must output ONLY a JSON array, do not output anything else, do not hallucinate or make up data, strictly follow the decision rules.

\end{mdframed}

\end{document}